\documentclass{article}

\usepackage[preprint]{neurips_2023}

\usepackage[utf8]{inputenc} 
\usepackage[T1]{fontenc}    
\usepackage{hyperref}       
\usepackage{url}            
\usepackage{booktabs}       
\usepackage{amsfonts}       
\usepackage{nicefrac}       
\usepackage{microtype}      
\usepackage{xcolor}         

\usepackage[disable]{todonotes}
\usepackage{multirow}
\usepackage{caption}
\usepackage{subcaption}
\usepackage{colortbl}
\usepackage{graphicx}
\usepackage{enumitem}

\usepackage{wrapfig}
\usepackage{multirow}

\usepackage{algpseudocode}
\usepackage{xcolor}
\usepackage[linesnumbered,ruled,vlined]{algorithm2e}

\SetCommentSty{mycommfont}

\SetKwInput{KwInput}{Input}                
\SetKwInput{KwOutput}{Output}              

\title{Quality Diversity in the Amorphous Fortress (QD-AF):\\ Evolving for Complexity in 0-Player Games}

\author{
    Sam Earle \\
    Game Innovation Lab\\
    New York University\\
    Brooklyn, NY\\
    \texttt{se2161@nyu.edu} \\
    \And
    M Charity \\
    Game Innovation Lab\\
    New York University\\
    Brooklyn, NY\\
    \texttt{mlc761@nyu.edu} \\
    \And
    Dipika Rajesh\\
    Independent Researcher\\
    Chennai, India\\
    \texttt{dipika.rajesh@gmail.com}
    \And
    Mayu Wilson\\
    Independent Researcher\\
    Brooklyn, NY\\
    \texttt{mayuwilson@gmail.com}
    \And
    Julian Togelius\\
    Game Innovation Lab\\
    New York University\\
    Brooklyn, NY\\
    \texttt{mlc761@nyu.edu} \\
}

\begin{document}

\maketitle

\begin{abstract}
We explore the generation of diverse environments using the Amorphous Fortress (AF) simulation framework. AF defines a set of Finite State Machine (FSM) nodes and edges that can be recombined to control the behavior of agents in the `fortress' grid-world. The behaviors and conditions of the agents within the framework are designed to capture the common building blocks of multi-agent artificial life and reinforcement learning environments. Using quality diversity evolutionary search, we generate diverse sets of environments. These environments exhibit certain types of complexity according to measures of agents' FSM architectures and activations, and collective behaviors. Our approach, Quality Diversity in Amorphous Fortress (QD-AF) generates families of 0-player games akin to simplistic ecological models, and we identify the emergence of both competitive and co-operative multi-agent and multi-species survival dynamics. We argue that these generated worlds can collectively serve as training and testing grounds for learning algorithms.
\end{abstract}

\section{Introduction}



Games with certain open-ended characteristics, such as sandbox simulation, management, or multiplayer games, provide promising testbeds for learning agents~\citep{earle2021video,fan2022minedojo,suarez2021neural}.
Such games allow for a range of potentially interesting interactions between systems and agents, often leading to emergent phenomena unforeseen by developers~\citep{guttenberg2023designing}.

\begin{figure}
    \centering
    \includegraphics[width=\linewidth]{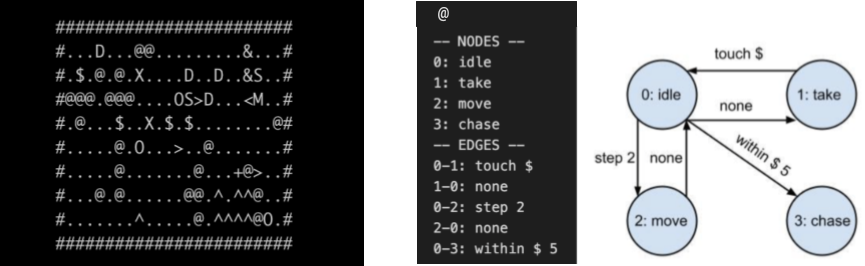}
    \caption{\textbf{Left: example of a random fortress} generated in the Amorphous Fortress framework. The fortress environment contains many instances of different entity class types. \textbf{Right: An example entity class FSM definition}. This FSM defines the action behavior an instance agent of this class can take within the fortress under the right conditions. Here, the agent is capable of random movement as well as `chasing' (moving along a path toward) and `taking' (removing from the fortress) an instance of the \$ entity class.}
    \label{fig:af-example}
\end{figure}

Amorphous Fortress~\citep{charity2023amorphous} is a simulation framework that uses finite-state machines (FSMs) to produce emergent AI behaviors.
Drawing inspiration from games such as \textit{Dwarf Fortress} and \textit{Rogue}, AF defines a base reality consisting of a fortress, where multiple instances of FSM agents interact with each other.
The agents in the fortress can perform abstract behaviors and interactions such as hunting, transforming, and reproduction, and these behaviors are triggered by temporal or spatial conditions dependent upon the agent's state (Figure~\ref{fig:af-example}).
We define a single fortress as containing a set of FSMs (one for each entity class), and a level topology (initial entity instances and coordinates).

Prior work discusses some of the non-trivial multi-agent dynamics that emerge when fortresses are optimized by a simple hill-climber algorithm for both the size of agent FSMs and the degree to which they are explored over the course of a simulation~\citep{charity2023amorphous}.
Fortresses are observed in which symbiotic relationships exist between entities; where entity classes with both large and small FSM graphs are integral to the fitness in the fortress.
Despite the simplicity of the hillclimber, a variety of entity classes emerge, with diverse policies of agent behavior that depend on one another for deeper exploration of their own graphs.

We extend this earlier work by optimizing both quality and diversity using metrics reflecting agent behavior and interaction.
We implement the quality diversity (QD) algorithm MAP-Elites to illuminate an archive of Amorphous Fortress fortress environments.
Diversity is maintained by grouping fortresses by behavior characteristics (BCs), and only having fortresses compete against others with similar BCs.
These BCs are measured based on the agent action space class definitions and the ending state of the fortress after simulation.
Quality of the fortresses is optimized by evaluating what proportion of each entity class's finite-state machine was explored (collectively, over all instances of the entity class) during simulation. We use these metrics to generate collections of fortresses that exhibit a range of specific behaviors during simulation.

\todo{Unsupervised approach for generating diverse and non-trivial environments}
QD-AF is an unsupervised approach for generating environments that are distinct from one another, with non-trivial and controllable simulation dynamics.
The diversity and complexity of environments generated by QD-AF makes them an ideal candidate for serving as a general testbed for evaluating RL agents.
\todo{more specifics about learning agent}
\todo{dedicated section: proposed learning agent. how to generate goals? Uses heuristics, static FSM analysis, or a PAIRED-like setup with a goal-generator? Look at population trajectories and select new target populations---more divergent from averages=more difficult}
\todo{mention hypothetical forward-predictor model?}

\section{Background and related work}

This extension of the Amorphous Fortress system emphasizes the themes of multi-agent open-ended environments and QD algorithms. The following subsections describes each theme in more detail along with previous works related to the experiments in this paper.

\subsection{Open-ended learning environments}

Simulation environments allow researchers to emulate real world events and phenomena in a controlled test framework. Open-ended simulation environments and environments that promote artificial life offer a multitude of challenges and emergent scenarios for AI to solve \citep{bedau2000open, stanley2017open}. Examples of simulation environments have previously been studied by game AI researchers for developing both the artificial agents and the environments themselves. 
\citet{charity2020say} and \citet{green2021exploring} introduce minimal simulations of The Sims and RollerCoaster Tycoon environments respectively to generate novel and diverse layouts based on the game environment. Similarly, \citet{earle2020using} introduces a training environment in the game SimCity and examines population behaviors of cellular automata in Conway's Game of Life (GoL).
To better understand the role of players in game design, \citet{bjork2012zero} discuss 0-player games, in which the player interacts with the game either in some highly constrained and atypical fashion or not at all. They describe GoL as a ``setup-only'' 0-player game, in which the player is only able to interact with the system prior to the simulation. In this light, QD-AF can be seen as optimizing the setup of such a 0-player game, with a QD algorithm used in place of a human player to search for mechanics and level topologies that produce interesting simulation rollouts in AF.

\textit{Minecraft} has been a popular environment for open-ended reinforcement learning tasks. \cite{fan2022minedojo} provide and train agents on a large set of RL tasks---some generated by humans, others the result of few-shot prompting an LLM for similar tasks. \citet{zhang2023omni} use human notions of interestingness in conjunction with Large Language Models to explore and facilitate open-ended learning. \cite{wang2023voyager} automatically prompt an LLM for new and interesting tasks given a player's progress, where the player is a similar LLM, interfacing with the game via a collection of programs defining player behavior, which it generates and may re-use over the course of play. In a parallel line of work focused on environment generation, Evocraft~\cite{grbic2021evocraft} is proposed as a framework for open-ended research in creating structures and systems with various functional or aesthetic ends using the game's set of blocks (which include musical note generators and``redstone'' circuits), as demonstrated by \cite{sudhakaran2021growing}. While AF is a much simpler framework than \textit{Minecraft}, its approach to evolving populations of FSMs could be adapted for the generation complex multi-agent dynamics in Minecraft. Conversely, AF could be extended to involve non-agentic objects (analogous to Minecraft blocks) serving as (possibly destructible) obstacles or resources.

Languages like Griddly~\citep{bamford2021griddly}, VGDL~\citep{schaul2013video}, and PuzzleScript~\citep{lavelle2013puzzlescript} provide domain-specific languages for describing game-like environments for reinforcement learning (RL), search and planning algorithms, or human play.
Such systems are sometimes accompanied by web frameworks allowing users to edit and upload their own maps~\citep{bamford2022griddlyjs}.
The trajectory of such databases over time (e.g. where RL agents learn on a growing number of human-designed maps) can be seen as human-in-the-loop open-ended artificial learning processes, reminiscent of PicBreeder~\citep{secretan2008picbreeder}, where users collaboratively evolved small networks encoding images.
Amorphous Fortress would make an ideal candidate framework for such an online open-ended loop, allowing users to assign meaning via recombination of abstract, atomic mechanics and map layouts.




 

 
\subsection{Multi-agent interactions}

In some open-ended environments, multiple decision-making agents interact with each other co-operatively or competitively to achieve their common or opposing objectives.
\citet{deshpande2021drawcto} introduce a web application---Drawcto---which uses multiple agents that are capable of co-creating interpretive open-ended artwork with human collaborators. Neural MMO~\citep{suarez2021neural}, is a platform for multi-agent RL over large agent populations in procedurally generated virtual maps. \citet{lowe2017multi} introduce a method that successfully learns RL policies that require complex multi-agent co-operation and coordination.
In the framework proposed by \citet{grbic2021evocraft}, an example is given in which a human user interacts with an evolutionary algorithm to co-operatively generate complex block structures inspired by Minecraft.

Moreover, the diverse interactions among these multiple agents could give rise to interesting emergent behavior within the given context. Work by \citet{bansal2018emergent} and \citet{baker2019emergent} introduces the simulation of diverse environments, where multiple agents engage with each other competitively, leading to the rise of intricate and complex emergent behaviors. Such emergent behaviors include offensive and defensive game playing strategies like blocking and kicking~\citep{bansal2018emergent} or object manipulation within the environment such as ``box-surfing'' and building barricades~\citep{baker2019emergent}. We employ multi-agent interactions to enhance the diverse and interesting emergent behavior of different entity class types within the fortress environment.



\subsection{Quality diversity algorithms}

Evolutionary algorithms are gradient-free optimization methods that randomly mutate pools of individuals to maximize a computable objective function~\citep{russell2010artificial}.
Novelty search, an extension of evolutionary algorithms, replaces the objective function with a measure of an individual sample's phenotypic/behavioral distance from the existing archive of discovered individuals, in effect uniform randomly sampling the search space~\citep{doncieux2019novelty}, often achieving the (held-out) objective given a sufficiently informative behavior distance metric.
Novely search has been used to generate diverse game AI components such as video game levels~\citep{beukman2022procedural} and dungeons~\citep{melotti2018evolving}. 
The novelty search algorithm has served as a precursor to another derivative of evolutionary algorithms known as Quality Diversity (QD) algorithms. These algorithnms are designed to maintain diversity while optimizing quality among a population of individuals.

The QD algorithm Multi-dimensional Archive of Phenotypic Elites (MAP-Elites)~\citep{mouret2015illuminating} optimizes for a fixed objective while tessellating a behavioral search space and preventing competition between elites in different cells.
MAP-Elites has been used to generate teams of agents for automated gameplaying~\citep{guerrero2021map}.
MAP-Elites has also been used to create, replicate and explore real-world adaptability by simulated agents in virtual open-ended environment as studied by \citet{norstein2022open}. \citet{pierrot2023evolving} evolve repertoires of full agents to combine any RL algorithm with MAP-Elites to dynamically learn the hyperparameters of the RL agent.
This approach not only alleviates the user's workload but also enhances performance in the evaluated environments. In this work, we use MAP-Elites to generate the multiple finite-state machine agents that would be ideal use case environments for training agent learning models.

\subsection{Unsupervised environment design}

Existing work in Unsupervised Environment Design (UED) uses feedback from RL player-agents to incentivize the generation of novel game environments (which is in turn cast as an evolutionary or RL optimization problem)~\citep{parker2022evolving,jiang2021prioritized,dennis2020emergent,wang2020enhanced}, with recent extension to multiplayer games~\citep{samvelyan2023maestro}.
In UED, the objective of the environment generation and curation processes is some proxy for the learnability or relevance of new environments to the RL player, and the player is found to be more robust when trained in such a setting as opposed to domain randomization.
Elsewhere, RL and QD have been applied to a variety of level-generation problems, where domain-specific heuristics or fixed, search-based agents can be used to condition the goals of the generator~\citep{khalifa2020pcgrl,earle2022illuminating}.
In QD-AF, environments are populated with FSM-based player agents, and these players and their environments (i.e. the initial fortress topologies) are jointly optimized for metrics of complexity based on agent interaction. Our hope is that these metrics will serve as meaningful proxies for the generation of an initial set of learning environments which can ultimately be used to jump-start a UED process with online RL player agents.


\section{Amorphous Fortress 1.0 framework}

\todo{mention 15 entity classes, 94 possible node types, 1410 in aggregate over entities...}
The Amorphous Fortress framework, developed by \citet{charity2023amorphous}, is an artificial life simulation system has a hierarchy of 3 components: entities (the agent class of the system) the fortress object (the environment class of the system) and the engine (the ``manager'' and main loop of the simulation). Each entity of the Amorphous Fortress is defined by a singular ASCII character, a unique 4-bit identification hex number and a finite-state machine (FSM) specifying its behavior during simulation. The finite-state machine entity class definition is made up of a list of nodes and a set of edges. Each node in the FSM graph represents a potential action state an entity instance can be in. These actions define how an instance interacts with the environment. The edges define when an instance of the entity class can change states to another node and is dependent and prioritized based on conditions found during simulation. Table \ref{tab:action_node} shows the possible action nodes and Table \ref{tab:condition_edge} shows the possible conditional edges that can define an entity class for the Amorphous Fortress 1.0 System. At any time during the simulation, the entity is always in a state at one of the set nodes. At each timestep---a single update within the fortress environment---each connection is evaluated to move to the connecting node state based on whether the conditions are met, in order of priority defined internally. The agent will perform the action at its new current node on the next timestep. 

The fortress of the Amorphous Fortress contains the environment where the simulation takes place and stores general information accessible to all of the entities in the fortress. The borders and size of the fortress are defined initially with a set character width $w$ and height $h$ to enclose the entities. On initialization, the fortress generates each entity class FSM for each character defined at the start of the simulation. This global dictionary of entity classes allows any instance of an entity to add or transform different entity instances even if none exist on the map at initialization. The fortress maintains a list of currently active entity instances in the simulation and adds or removes them by their ID value. The fortress also maintains positional data about each entity to return for conditional checks (i.e. whether a particular position has an instance of an entity class located there.) The engine of the Amorphous Fortress system maintains the execution of the simulation and exports any log files or entity class data information such as the node and edges definitions.
In this experiment, the fortress area itself where the entities interact is 15 spaces wide by 8 spaces tall---including the walls---allowing for a total traversable area of 78 tiles. Each fortress contains 15 entity classes, where each class can have a minimum of 1 action node and a maximum of 94 action nodes. 

As an update from this first Amorphous Fortress framework, the \textit{push} action node was modified to include an entity character target. The \textit{move\_wall} action node is also an addition from the previous iteration of the Amorphous Fortress system. 



\section{Methods}

\todo{Write out fortress mathematically in methods section}

\subsection{MAP-Elites for Amorphous Fortress} 
\begin{figure*}
    \centering
    \includegraphics[width=0.98\textwidth]{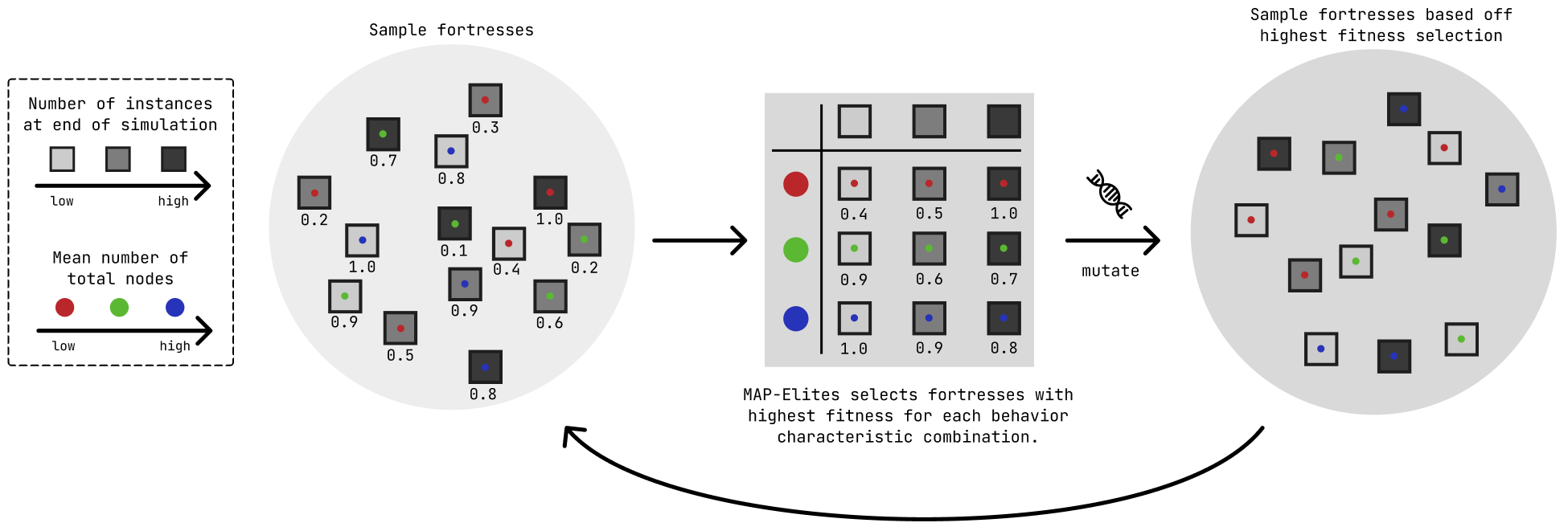}
    \caption{QD-AF uses MAP-Elites to iterate on an archive of elites differentiated in terms of the size of their FSMs, and the mean number of surviving instances in the fortress after simulation. Fortresses in the same part of the archive compete to maximize the overall proportion of agent FSMs explored during simulation.}
    \label{fig:map-elites}
\end{figure*}

We implement the MAP-Elites QD algorithm \citep{mouret2015illuminating} to evolve multiple entity classes towards a diversity of emergent behaviors. 
MAP-Elites evaluates a population of samples based on a fitness function while maintaining a grid of these samples and organizing them based on pre-defined Behavior Characteristics (BCs) exhibited during evaluation to maintain both quality and diversity of generated individuals. 
The emergent behaviors of the entities defined within this system are dependent upon interactions with other instances within the same fortress. Therefore a single cell of the MAP-Elites grid contains a fortress with a set of entity class definitions. Figure \ref{fig:map-elites} illustrates a small example of the evolution and evaluation process as the fortresses are placed in the MAP-Elites archive for the experiment (described in more detail later in this section).

\paragraph{Behavior Characteristics}
\todo{use $n$ instead of 15}
We use the following BCs: a) the mean number of total instances in the fortress at the end of the simulation and b) the mean number of total nodes across all entity class definitions.

Behavior characteristic (a), based on the average number of entity instances surviving in the fortress over the course of a set of simulation episodes (with different random seeds), is intended to explore the population dynamics within fortresses.
\todo{maybe move this to results}
That is, we seek fortresses in which the entity class combinations result in an overpopulation of entity instances, an extinction of all instances, or a stability or ``equilibrium'' of instances within the fortress. The values of this dimension can range from 0 to 156---the maximum number of instances allowed to exist in the fortress before it terminates based on an ``overpopulation'' condition. 

Behavior characteristic (b), based on the collective class FSM size, looks to examine the complexity and depth of the entity classes; whether the combined set includes a majority of simple entity class definitions with only 1 action node or conversely with extremely large entity class definitions. The values of this dimension can range from 15 nodes---where each of the entity classes has only 1 node---to 1410 nodes ($max\_num\_nodes = [(6 \times n) + 4] \times n$ where $n$ is the number of entity classes in the fortress, with $n=15$ in our experiments)---where each possible node is included every entity class FSM definition. The exploration of this dimension by the MAP-Elites algorithm will demonstrate the growth and utility of varying sized entity class FSMs. 

\paragraph{Mutation}
\todo{Be more specific about FSM and map mutation, harmonic distributions and so on}
A fortress individual in the population is mutated by modifying its genotype: the class level definitions of the FSMs. \todo{maybe remove this comparison to earlier work} As in \citet{charity2023amorphous}, where separate coin-flip probabilities determine whether a node, edge, and/or entity instance in the fortress itself is added, removed, or altered. Here, the node mutation is adjusted to increase exploration within the grid. A random range of nodes can be added, removed, or altered into another action node definition during a single step of mutation. For example, 10 nodes can be added to one entity class definition, while 4 are removed from another (or the same if randomly chosen again). Algorithm \ref{alg:mutation} shows a pseudocode algorithm for the mutation process of the evolution.

\todo{trim this, still mention mutation/new-random individuals}
In our experiments, for every $9$ mutated individuals, a random fortress is added to the offspring pool to encourage exploration within the MAP-Elites grid and to prevent the algorithm from reaching a local minimum during evolution. 
On initialization, all 15 of the entity class FSMs are randomly generated for each fortress. 

We randomly initialize fortresses so as to sample uniformly along the axis measuring number of aggregate FSM nodes. We first uniformly sample this aggregate number, then split it into as many summands as there are entity types using an evenly weighted multinomial distribution, where each summand corresponds to the number of nodes to be assigned a given entity. It is possible in this setting for an entity type to be assigned more nodes than there are distinct node types; in this case, we (greedily) re-assign the surplus nodes to one or more non-overfilled entity classes, until no surplus nodes remain.

\paragraph{Fitness}

The fortress is simulated for 100 steps, where each instance of an entity class present in the fortress enacts the current action node of its FSM graph once per step and then evaluates the next action node to move to based on the state conditions it ends in. 
The fitness $f\in [0,1]$ of a fortress is given by $f = e / t$.
$t$ is the total number of nodes and edges in the set of entity FSMs, and $e$ counts the number of explored nodes and edges across entity FSMs---those that were activated at least once over the course of simulation.
The fitness function is similar to the hill-climber experiment from the original Amorphous Fortress work, which is based on the average proportion of nodes and edges that have been explored, i.e., the percentage of nodes over the whole entity class activated during simulation by all instances of said class.
\todo[inline]{cut this sentence or split into 2}
This fitness definition encourages each class entity to explore the full possibility of its emergent behaviors demonstrated within the simulation. The final exploration of a class definition's nodes and edges are also aggregated over evaluation trials in case different behaviors occur due to different seed evaluations. 
\todo[inline]{specify fitness---FSM activation}

\paragraph{Entropy of FSM definitions}

Given an archive of fortresses---optimized to maximize FSM exploration while diversifying along number of surviving entity instances and number of total FSM nodes---we additionally investigate the distribution of total FSM nodes among entity classes. More specifically, we define a handful of FSM size buckets, and measure the entropy of the distribution of entity class FSMs among these buckets (Figure~\ref{fig:exp1_entropy_archive}). The entropy of the FSM size distribution is necessarily minimal at the extremes along the axis measuring number of nodes: only a set of minimum/maximum size FSMs can sum to a minimum/maximum number of total FSM nodes, where in either case, all FSMs must have the same size (falling into to the same FSM size bucket and minimizing distributional entropy). 
The values of this dimension range from 0 to 1, with 0 meaning all of the entity class FSMs have the same number of nodes and no variation, and 1 meaning the number of nodes are different for each entity class FSM definition. We use Shannon Entropy to calculate the entropic value of the FSM sizes with a $b$ base $N$ where $N$ is the number of FSM size bins (which for this experiment is equal to the number of entity classes).
\todo[inline]{set number of size buckets the same as entity classes}
\todo[inline]{maybe describe that the shannon entropy represents bits of info needed to encode a thing}
\todo[inline]{the base change is an implementation detail that we use to get our values between 0 and 1}

\begin{figure*}[!ht]
    \centering
    \includegraphics[width=0.98\textwidth]{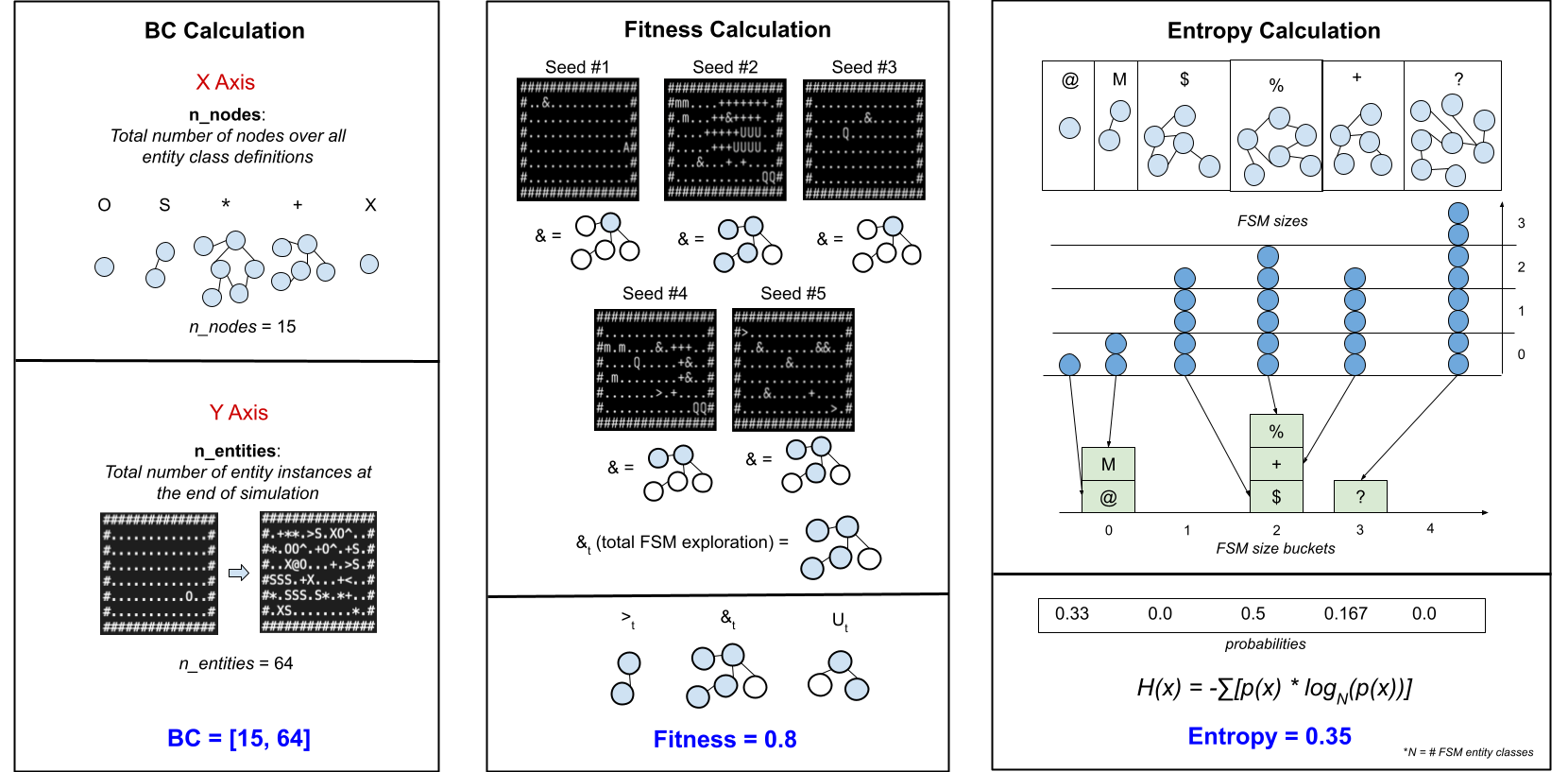}
    \caption{Diagram of the calculations for MAP-Elites BCs, fitness values, and entropy value. Our BCs involve counting the number }
    \label{fig:map-elites_bc-f-e}
\end{figure*}

Figure \ref{fig:map-elites_bc-f-e} shows how the BCs, fitness value, and entropy value are calculated for any given fortress.
 
\label{sec:exp_setup}
\subsection{Experiment setup}

Using the fitness function, AF mutation operators, and AF initialization schemes described above, we use MAP-Elites to iterate on an archive of AF samples, evaluating them to determine fitness measures and characteristics, sorting them into their respective MAP-Elites grid cells, re-sampling from the archive to create the new population of samples and repeating for a set number of generations.
We set the number of entity classes to 15, evaluate these fortresses across 5 randomly chosen seeds, and evolve the populations for 10,000 generations.


\subsection{Considerations for injecting learning agents into the fortress}

We envision fortresses as potential training grounds for learning agents.
Though the joint optimization of entities and maps toward complex 0-player rollouts is worth studying on its own merits---and in one sense, the agents in QD-AF might be seen as learning (via evolution) to cooperatively maximize complexity metrics---we believe the fortress environment would be an ideal testbed for Multi-Agent Reinforcement Learning.

As an intermediary step, we can imagine training a predictive model or binary classifier to generate or identify possible completions in a series of frames from a sequence rollout.
We expect that fortresses exhibiting more complex behavior could take more time and/or network capacity to learn with high precision.
These discrepancies could be used as a more general and principled complexity metric---albeit one that is very expensive to compute---which could be used to validate the cheaper, domain specific FSM-based complexity measures explored here.
Such a measure of neural predictability or compressibility would also likely be indicative of the level of challenge a particular fortress would present to an RL agent assigned to control one or a set of agents within the fortress, in an embodied fashion.
This is because an agent having to deal with the activity within the fortress will most likely have to (at least partially) model the behavior of the other entities within it.

To inject an RL agent into a fortress, we would perform an analysis of its FSMs and of the results of several simulated rollouts to generate plausible tasks for the agent.
Selecting one entity type over which to give the RL agent control, we would allow the RL controller determine the low-level movement/directional actions that would normally be automatically generated by move, chase, and push nodes, whenever these nodes are active in an entity instance's FSM.
We would still allow non-navigational nodes such as `transform' or `take' to become active should the relevant edges be triggered, so that the original action space of the entity remains unchanged.
We would take note of entity populations showing variance over the course of multiple rollouts, the implication being that the RL controller should be able to affect the ultimate population size by executing careful low-level movement.
We could then reward the RL agent for simulations that approach certain population targets, with targets further away from the mean population values during random rollouts constituting more challenging tasks.


\section{Results}

\begin{figure*}[!ht]
\begin{subfigure}{\linewidth}
    \centering
    \includegraphics[width=\linewidth]{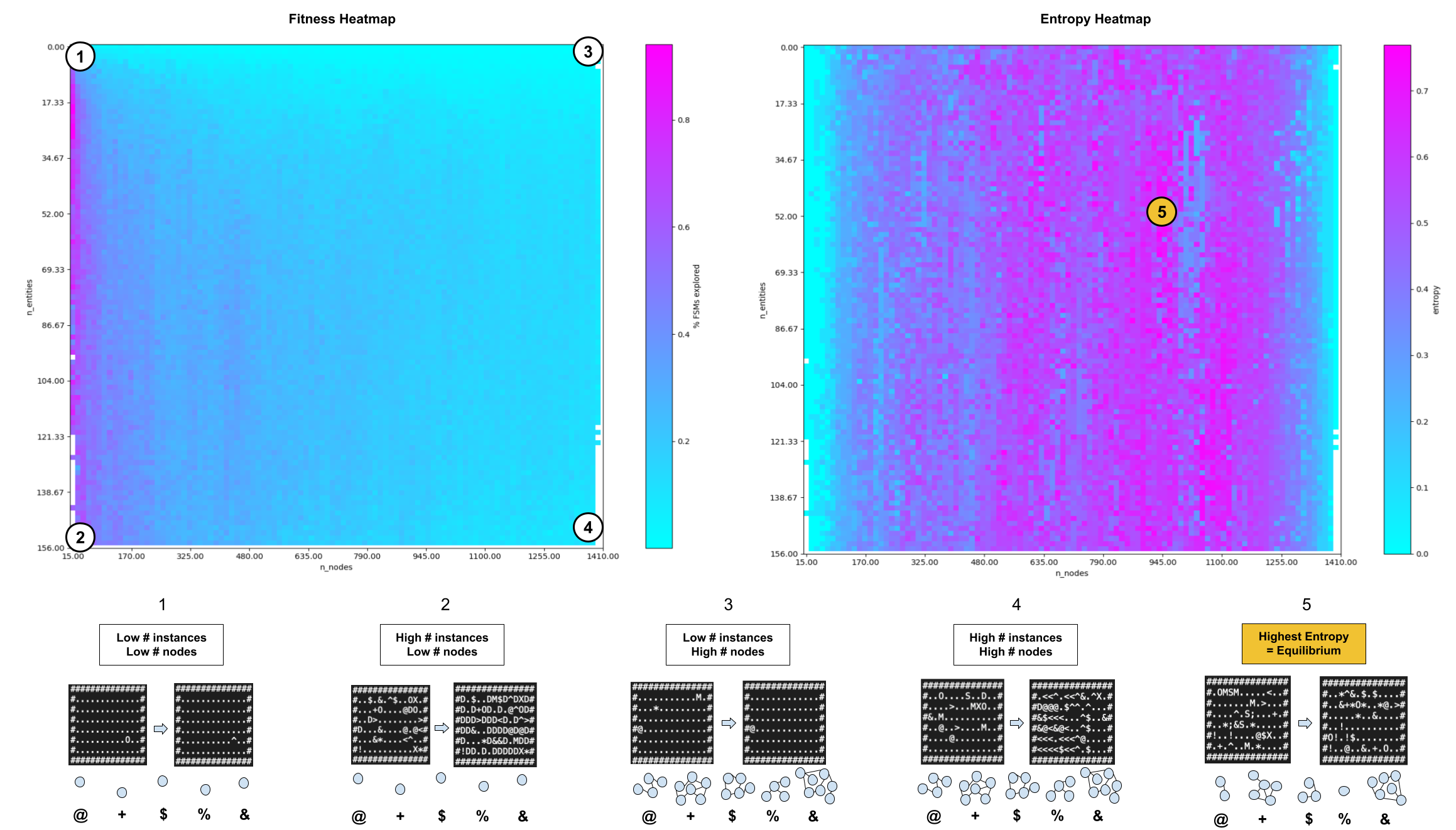}
    \caption{Archive of fortresses discovered by QD. Behavior characteristics are the \textbf{number of instances} at the end of the simulation vs. \textbf{total number of nodes}. The graph on the left shows the heatmap with respect to the fitness measurement---the proportion of FSMs explored in the fortresses. Naturally, FSMs with fewer nodes (left) are more easily explored. The graph on the right shows the same archive of fortresses, with heat denoting the entropy of FSM sizes within each fortress.}
    \label{fig:exp1_graphs}
\end{subfigure}
\begin{subfigure}{\linewidth}
    \centering
    \includegraphics[width=\linewidth]{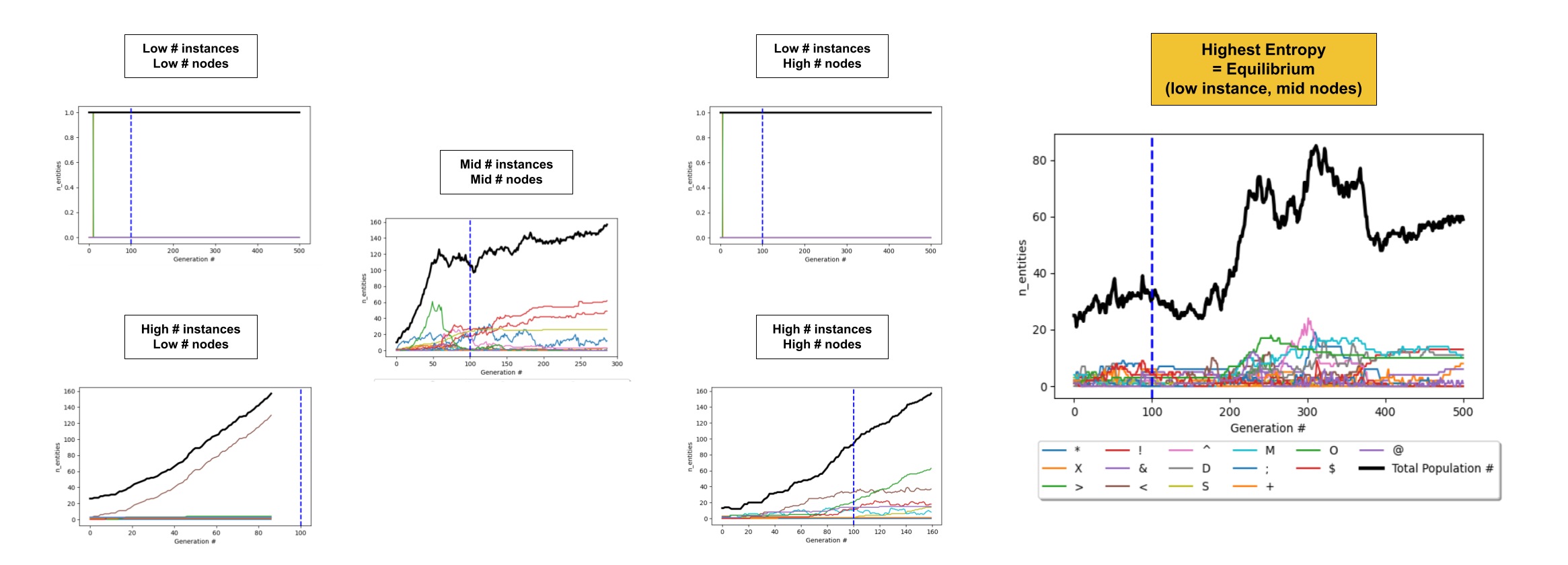}
    \caption{\textbf{Intra-fortress population}. For select fortresses in the archive, we measure the number of instances over time, across entity types. The vertical dashed blue line is the number of iterations the fortress was simulated for during evolution. The cell with the highest entropy of entity class definitions achieves an ecological equilibrium. This behavior is distinct from the extremes of the archive, which either overpopulate or rarely add any additional entities.}
    \label{fig:pop_growth}
\end{subfigure}
\caption{Exploring complexity of agent FSMs and inter-agent-type population dynamics in archives of fortresses evolved to optimize FSM exploration (avoiding unneeded nodes/edges) while exploring along axes of mean FSM size and overall agent population after simulation.}
\end{figure*}

Figure~\ref{fig:exp1_fit_archive} a heatmap of fitness over individuals in the MAP-Elites archive (where color indicates fitness).
Nearly every possible grid cell is filled for this experiment. Fortresses with fewer total nodes have the highest fitness values in the grid---most likely because it becomes increasingly challenging to explore different nodes in larger FSM graphs. Predictably, many fortresses recording a large number of surviving entity instances at the end of an episode were terminated early due to overpopulation. The number of total surviving entity instances does not seem to be limited by the total number of nodes, with fortress environments with varying FSM sizes leading to either extremes of near extinction (0-1 instances) or overpopulation (156 instances) and various states of equilibrium in between.

\paragraph{Measuring entropy}

Higher entropy FSM size distributions appear in the subsection of the axis containing fortresses with medium-high numbers of nodes. It is possible that more variably-sized sets of FSMs could be beneficial for maximizing fitness (in terms of FSM exploration during simulation), which might explain their prominence in this section of the archive. This prominence is striking given the evenly-weighted multinomial distribution used to distribute FSM nodes among entity classes from a fixed number of aggregate number of FSM nodes, which normally lead to low-entropy FSM size distributions (because each FSM is likely to be assigned roughly equal numbers of nodes). On the other hand, fitness in this section of the archive is low, such that it may be unlikely very much such selection pressure has been applied, making further analysis necessary to come to form conclusions along these lines.

\todo{to add: do we avoid degenerate high-fitness cases (?)}






\section{Discussion}

\todo{set this up in the intro}
Our results show promise in generating a diverse set of environments for learning agents.
Our fitness function emphasizes environments in which entities exhibit a diversity of behaviors (i.e. explore as many nodes in their constituent FSMs as possible).
By combining this objective with behavior characteristics measuring the overall size of FSMs (via number of nodes), we seek to generate a set of environments that may act as a curriculum for a future embodied learning agent, which would have to navigate and perhaps (indirectly) model the varyingly complex behavioral policies governing the activity of NPCs in the fortress. 


After the QD search process, we additionally evaluate the diversity of the entity classes within each fortress, measured as the entropy over the distribution of entity class FSM sizes. We note that high entropy---exhibited in a large swath of the archive with a medium-high number of nodes---corresponds to sets of entities with variably sized FSMs. When such individuals are fit (and FSMs have few ineffective nodes/edges), we have some indication that different types of entities will exhibit diverse behavior.
In this case, a hypothetical learning agent will be forced to adapt to a diversity of behavior profiles, increasing the richness of its task.
\todo{maybe consider degenerate cases? Can we argue that we think this alternate explanation isn't likely}

We observed an interesting phenomena within the MAP-Elites cells concerning the population numbers of each of the entities. Ideally, we were aiming to find fortresses with balanced interactions between entities. The cells found at the extreme points of the archive (i.e. cells with lowest and highest possible behavior characteristics) exhibited uncooperative behavior between the instances. Fortresses with lower instance numbers refused to populate and caused a stagnation in the fortress. Conversely, fortresses with higher instance numbers quickly overpopulated. However, fortress individuals found in the middle of the archive had more ``equilibrium'' and cooperation. The fortress in the exact middle of the archive achieved much more diversity in terms of the entity class population; some entity classes having a sudden growth in instance number before dying off, while others slowly expanded their presence over time. This ``equilibrium'' was most noticeable in the fortress individual that demonstrated the highest entropy between entity class FSM sizes. This fortress showed a near perfect balance between all entity classes; neither dominating nor diminishing in numbers. The entities found in this fortress find a ``harmony'' of co-existence where the ecosystem does not find itself in danger of overpopulation nor extinction. From this, we can conclude that having a diversity of entity class sizes leads to better balance of entity populations and allows for more exploration of co-operative class behaviors.

The main weakness of our results is the generally low fitness, which indicates that much of the larger FSMs generated by our system could be pruned to drastically smaller size without having any effect on environment dynamics. We hypothesize that this lack of FSM exploration is the result of limited compute resources. In particular, $100$ steps of simulation is not likely enough to explore FSMs with up to $94$ nodes. Or, the small map size of generated environments may make prohibit more interesting large-scale dynamics.
We observe the QD score to still be rising steadily after 10k iterations, such that further evolution would be beneficial. 
Qualitatively, we see that certain fortresses in the archive maintain varying equilibria between entity types over long time horizons, potentially showing how to optimize for environments facilitating novel dynamics for lifelong learning agents.



\section{Future work}

From the engineering side, the fortress engine could likely be drastically accelerated if it was implemented in a batched, GPU-compatible manner, similar to the recent trend in RL environments which has allowed for orders of magnitude increases in simulation speed~\citep{lange2023evosax,freeman2021brax}.

Future work could explore adding BCs to control specific aspects of agent behavior.
For example, one archive dimension could measure how many times the ``take'' node is enacted by entities could encourage the evolution of fortresses with more or less aggressive entities. We would also like to examine the compressibility (e.g. via a simple gzip algorithm) or predictability (e.g. by a a neural network trained with supervised learning) of environment rollouts generated by a given fortress definition. We expect that such measures will provide a reasonable estimate of a hypothetical learning agent's ability to model and/or adapt its behavior to a given fortress \citep{gomez2009measuring}, and could thus be used to validate the effectiveness of our FSM-based complexity metrics (themselves being cheaper to compute) and/or supplant them (if necessary).

A different line of future work---emphasizing the QD-AF paradigm as a design tool in its own right---will involve developing a mixed-initiative online system in which users are free to design their own fortresses and entity class definitions.
The MAP-Elites fortress illumination process could create ``casts'' of generated characters for the user to include, acting as a recommendation engine.
These generated entity classes would be selected to highlight and enhance the potential behaviors singular ``main character'' entity within the fortress (e.g. by leading to particular activity in the main character's FSM). 
Future systems could then augment QD-AF with learned models of human preference and style gathered consensually via such an interface.
Users could even be invited to narrativize the emergent dynamics of fortresses in natural language, opening the door for training models converting human narrative and first-hand experience into environments---with real stakes and incentives beyond next-token prediction---for learning agents.

\section{Conclusion}

We use quality diversity evolutionary search to create an archive of grid-world environments using the mechanics of Amorphous Fortress, with an eye toward generating diverse training sets for learning agents.
By searching for diverse fortresses in terms of number of surviving entities at the end of a simulation, we guarantee that a hypothetical learning agent will be exposed to a variety of environment states.
In the archives generated by QD search, we find a large swath of environments which avoid extinction or population explosion to maintain equilibria that appear robust to stochasticity and longer episode lengths.
We select for fortresses with well-explored FSMs to prohibit the growth of ineffective FSM components. By diversifying the aggregate size of entity FSMs within an individual fortress, we seek to provide a set of environments containing a smooth increase in the complexity of agent behavior profiles.




\section{Acknowledgements}
We would like to thank Daphne Cornelisse, Noelle Law, Lisa Soros, and Graham Todd for their valuable input and feedback on drafts of this paper.



\bibliographystyle{ACM-Reference-Format} 
\bibliography{ref}

\newpage
\appendix


\begin{table}[!ht]
    \caption{Entity FSM action node definitions}
    \label{tab:action_node}
    \begin{tabular}{p{0.15\linewidth}  p{0.8\linewidth}}
    \toprule
        Action Node   & Definition                  \\
        \midrule
        idle          & \textit{the entity remains stationary atfferently-sizedsame position}  \\
        move          & \textit{the item moves in a random direction (north, south, east, or west)}              \\
        die           & \textit{the entity is deleted from the fortress}      \\
        clone         & \textit{the entity creates another instance of its own class}    \\
        push (c)      & \textit{the entity will attempt to move in a random direction and will push an entity of the specified target character into the next space over (if possible)}  \\
        take (c)      & \textit{the entity removes the nearest entity of the specified target character}     \\
        chase (c)     & \textit{the entity will move towards the position of the nearest entity of the specified target character}     \\
        add (c)       & \textit{the entity creates another instance from the class of the specified target character}      \\
        transform (c) & \textit{the entity will change classes altogether to an entirely different entity class - thus changing its FSM definition entirely}        \\
        move\_wall (c)      & \textit{the entity will attempt to move in a random direction unless there is an entity of the specified class at that position - otherwise it will remain idle}
    \end{tabular}
\end{table}

\begin{table}[!ht]
    \caption{Entity FSM conditional edge definitions (ordered by least to greatest priority)}
    \label{tab:condition_edge}
    \begin{tabular}{p{0.25\linewidth}  p{0.75\linewidth}}
    \toprule
        Action Node   & Definition                  \\
        \midrule
        none          & \textit{no condition is required to transition states}  \\
        step (int)    & \textit{every x number of simulation ticks the edge is activated and the node transitions}              \\
        within (char) (int)        & \textit{checks whether the entity is within a number of spaces from an instance of another entity with the target character}      \\
        nextTo (char)        & \textit{checks whether the entity is within one space (north, south, east, or west) of another entity of the target character}    \\
        touch (char)      & \textit{checks whether the entity is in the same space as another entity of the target character}  \\
    \end{tabular}

\end{table}


\begin{algorithm}[ht!]
\caption{Mutation function for the Fortress}\label{alg:mutation}
\KwInput{$node\_prob$, $edge\_prob$, $instance\_prob$}

$node\_r$ = random()\;
$edge\_r$ = random()\;
$instance\_r$ = random()\;

\tcc{Mutate random entity class nodes}
\While{$node\_r < node\_prob$}{
    $i$ = random(0,2)\;
    $e$ = random($fortress.ent\_def$)\;
    $n$ = random(log\_f())\;
    \uIf{$i == 0$}{
        $fortress.\_delete\_nodes(e,n)$\;
    }\uElseIf{$i == 1$}{
        $fortress.\_add\_nodes(e,n)$\;
    }\uElseIf{$i == 2$}{
        $fortress.\_alter\_nodes(e,n)$\;
    }
    $node\_r$ = random()\;
}
\tcc{Mutate random entity class edges}
\While{$edge\_r < edge\_prob$}{
    $i$ = random(0,2)\;
    $e$ = random($fortress.ent\_def$)\;
    \uIf{$i == 0$}{
        $fortress.\_delete\_edge(e)$\;
    }\uElseIf{$i == 1$}{
        $fortress.\_add\_edge(e)$\;
    }\uElseIf{$i == 2$}{
        $fortress.\_alter\_edge(e)$\;
    }
    $edge\_r$ = random()\;
}
\tcc{Mutate random entity instances in the fortress}
\While{$instance\_r < instance\_prob$}{
    $i$ = random(0,1)\; 
    $e$ = random($fortress.entities$)\;
    \uIf{$i == 0$}{
        $fortress.\_remove\_entity(e)$\;
    }\uElseIf{$i == 1$}{
        $x,y$ = random(fortress.pos)\;
        $fortress.\_add\_entity(e,x,y)$\;
    }
    $instance\_r$ = random()\;
}
\label{alg:qdaf}
\end{algorithm}


\begin{figure*}
\begin{subfigure}{\textwidth}
    \centering
    \includegraphics[width=.7\linewidth]{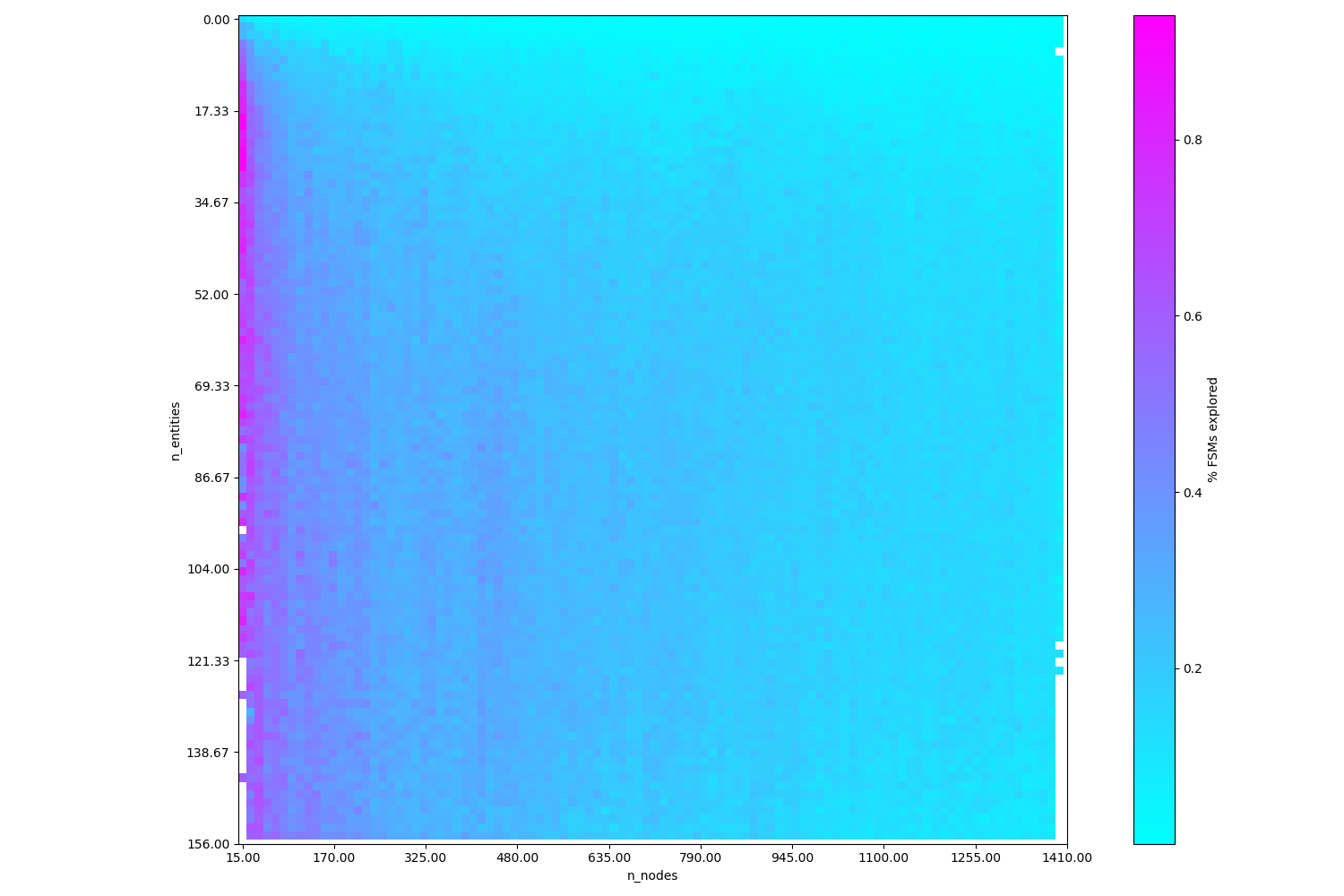}
    \caption{Proportion of FSMs explored in fortresses. Naturally, FSMs with fewer nodes (left) are more easily explored.}
    \label{fig:exp1_fit_archive}
\end{subfigure}
\begin{subfigure}{\textwidth}
    \centering
    \includegraphics[width=.7\linewidth]{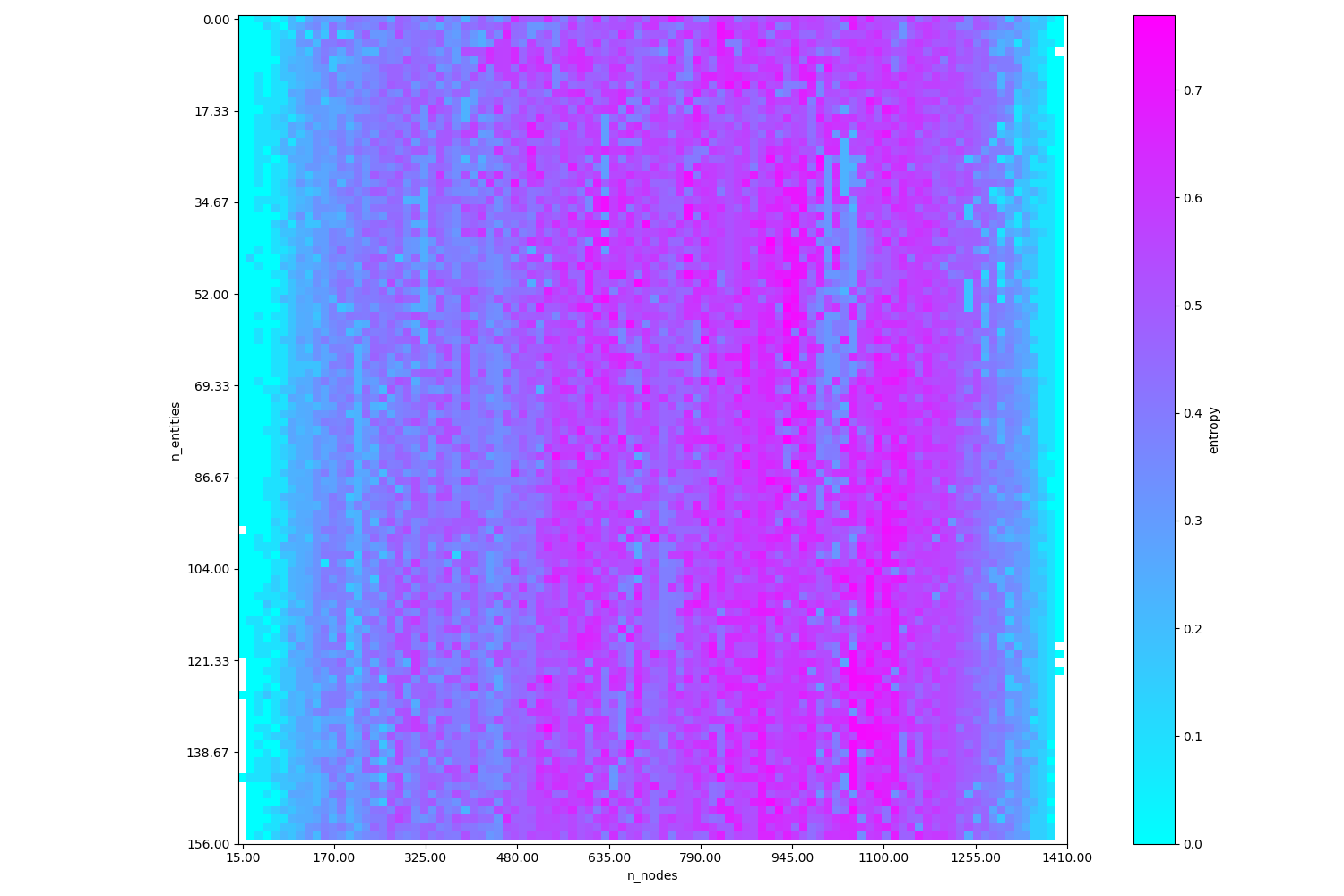}
    \caption{Entropy of distribution of FSM sizes across entity types. Fortresses with very few/many nodes (left/right) across all FSMs must have low entry because all entity FSMs are necessarily small/large. Fortresses with a medium-high number of nodes---allowing for diverse FSM sizes between entities---exhibit high entropy. This suggests that sets of differently-sized FSMs are more likely to result in thorough FSM exploration.}
    \label{fig:exp1_entropy_archive}
\end{subfigure}
\caption{Archive of fortresses resulting from maximizing proportion of FSMs explored while maintaining diversity in terms total size of FSMs and number of entity instances present in the fortress at the end of simulation.}
\end{figure*}

\begin{figure*}
\begin{subfigure}{\textwidth}
    \centering
    \includegraphics[width=.7\linewidth]{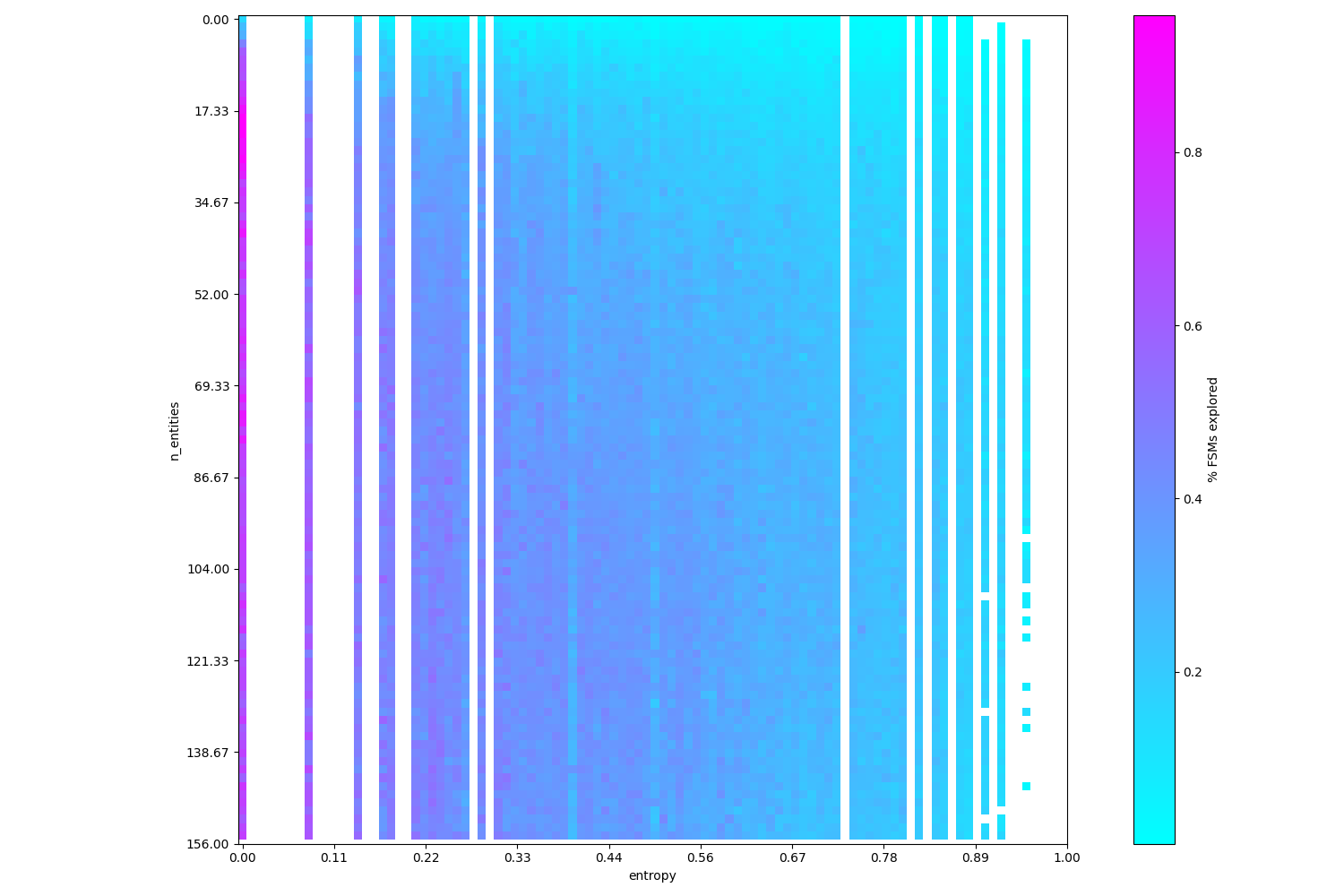}
    \caption{Proportion of FSMs explored in fortresses. Low entropy fortresses (left)---in which all entities have similar FSM size---allow for the most thorough exploration.}
    \label{fig:exp2_archive_fit}
\end{subfigure}
\begin{subfigure}{\textwidth}
    \centering
    \includegraphics[width=.7\linewidth]{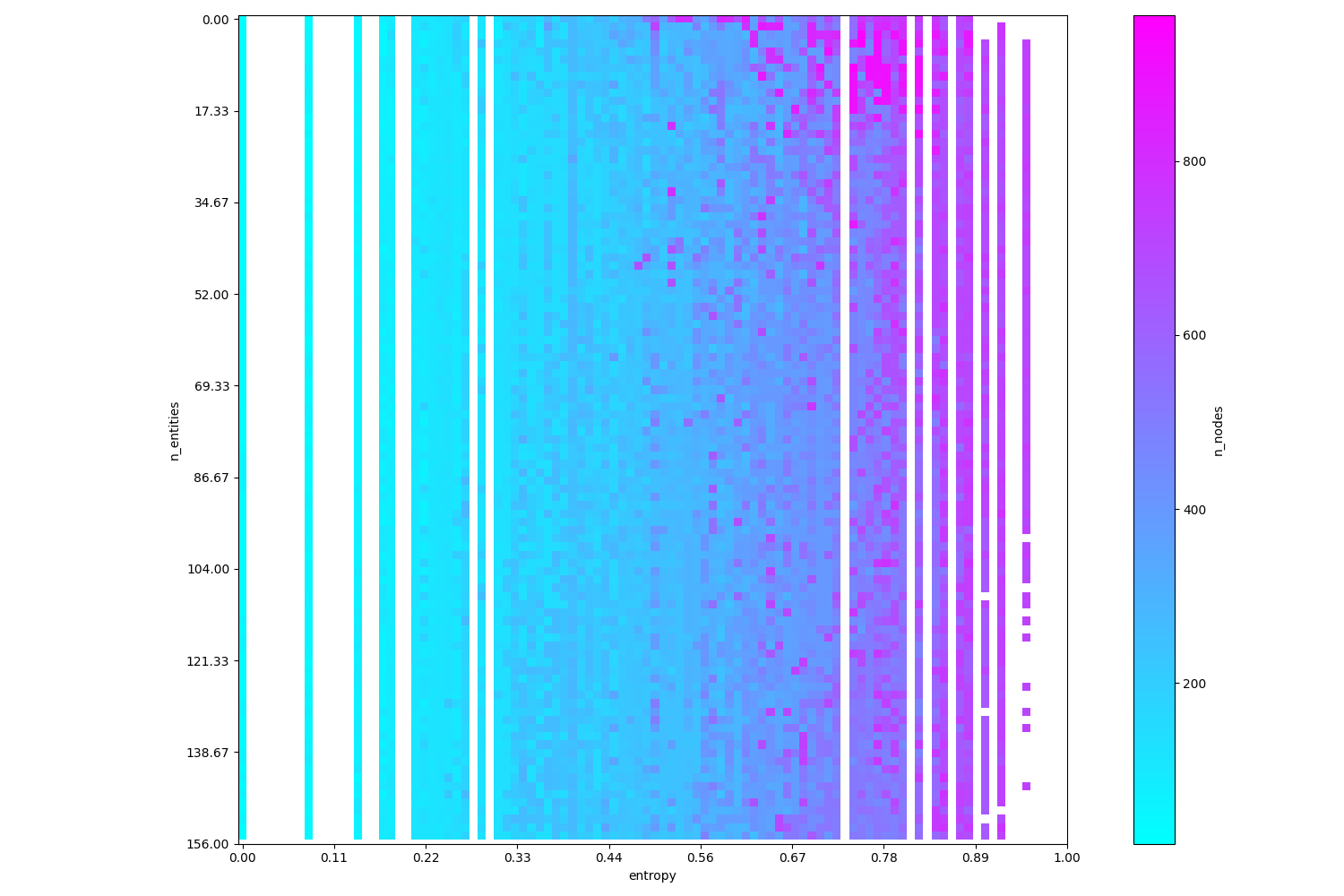}
    \caption{Number of nodes over all entity types. Naturally, minimal FSMs lead to the fittest low-entropy fortresses (left), while higher entropy FSM size distributions require more nodes overall (right). }
    \label{fig:exp2_archive_n_nodes}
\end{subfigure}
\caption{Archive of fortresses resulting from maximizing proportion of FSMs explored while maintaining diversity in terms of number of entity instances present in the fortress at the end of simulation, and entropy of the distribution of FSM sizes across entity types.}
\end{figure*}

\begin{table*}
\begin{center}
\begin{tabular}{||c c c c c c||} 
 \hline
 behavior characteristics & new seeds & n. episode steps & best score & QD score & archive size\\ [0.5ex] 
 \hline\hline
\multirow{3}{10em}{n. entities,\\n. nodes} & no & 100 & 0.941 & 2,235 & 9,986 \\ 
 & \multirow{2}{1em}{yes} & 100 & 0.941 & 2,197 & 9,975 \\ 
&  & 500 & 0.941 & 2,086 & 9,962\\ 
 \hline\hline
 \multirow{3}{10em}{n. entities,\\FSM size entropy}& no & 100 & 0.958 & 2,156 & 6,974 \\ 
 & \multirow{2}{1em}{yes} & 100 & 0.958 & 2,005 & 6,951 \\ 
 &  & 500 & 0.958 & 2,011 & 6,950 \\ 
 \hline
\end{tabular}
\caption{\textbf{Re-evaluation of elites with new random seeds and longer episodes.} After aggregating (re-evaluated) elites from 10 trials, we see that the stochastic nature of our environment leads to some variance, with some shrinking of the archive and decrease in QD score.}
\end{center}
\end{table*}

\begin{figure}[!ht]
\centering
\begin{subfigure}[t]{0.32\linewidth}
\includegraphics[width=\linewidth,trim={170 40 100 50},clip]{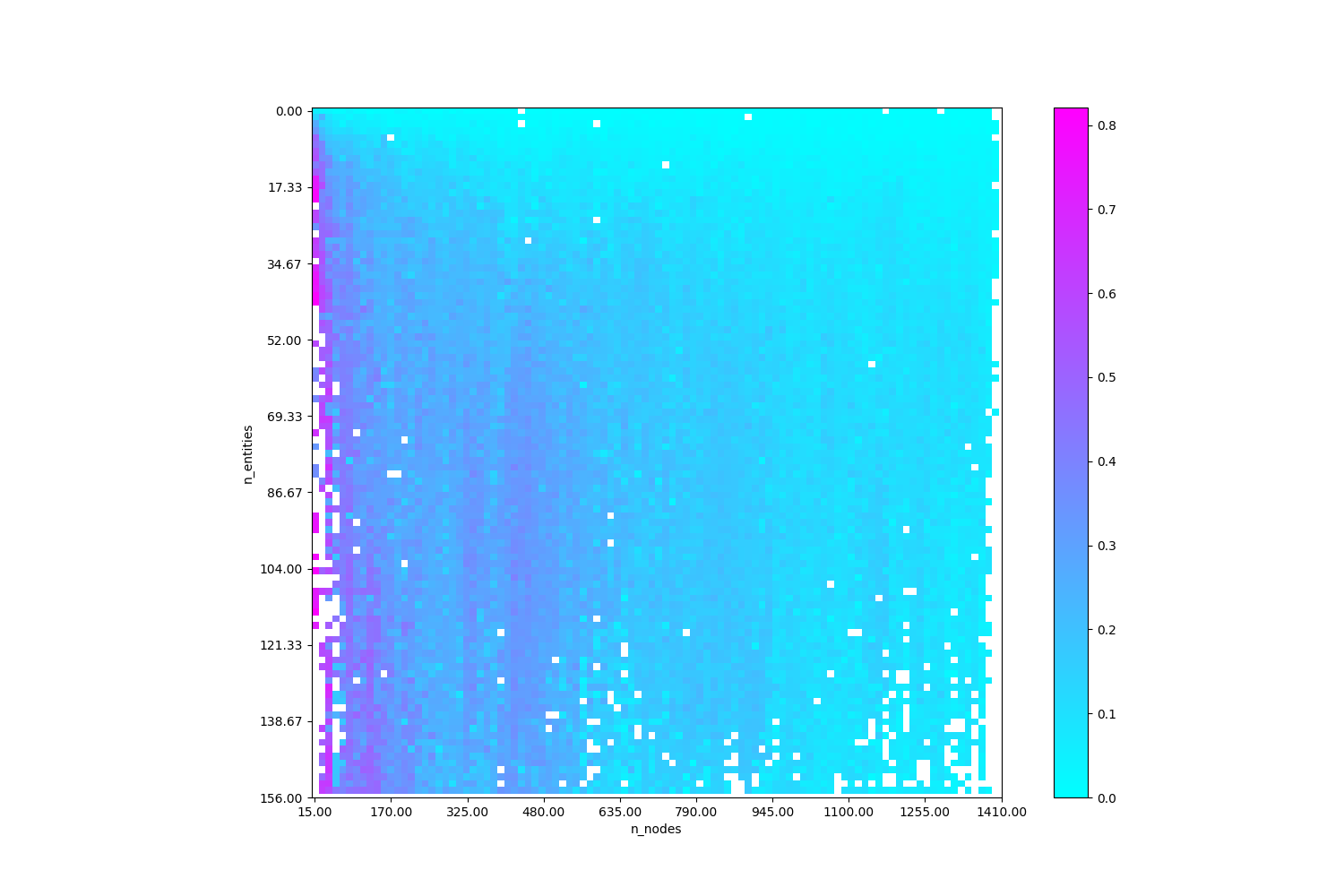}
\caption{Original archive after QD search.}
\label{fig:exp1_trial20_archive}
\end{subfigure}
\hfill
\begin{subfigure}[t]{0.32\linewidth}
\includegraphics[width=\linewidth,trim={170 40 100 50},clip]{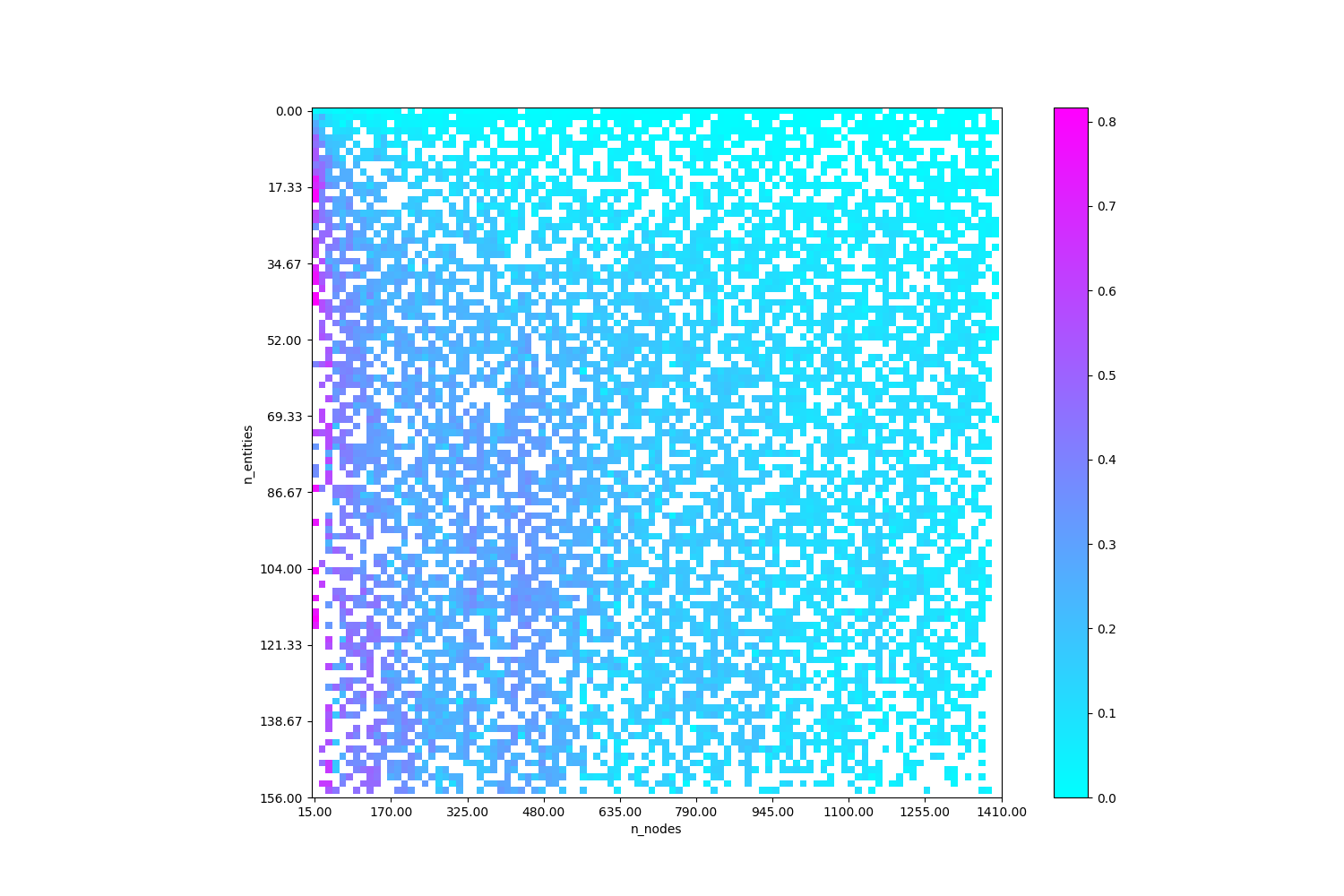}
\caption{Archive after re-evaluation with new random seeds.}
\label{fig:exp1_trial20_swiss}
\end{subfigure}
\hfill
\begin{subfigure}[t]{0.32\linewidth}
\includegraphics[width=\linewidth,trim={170 40 100 50},clip]{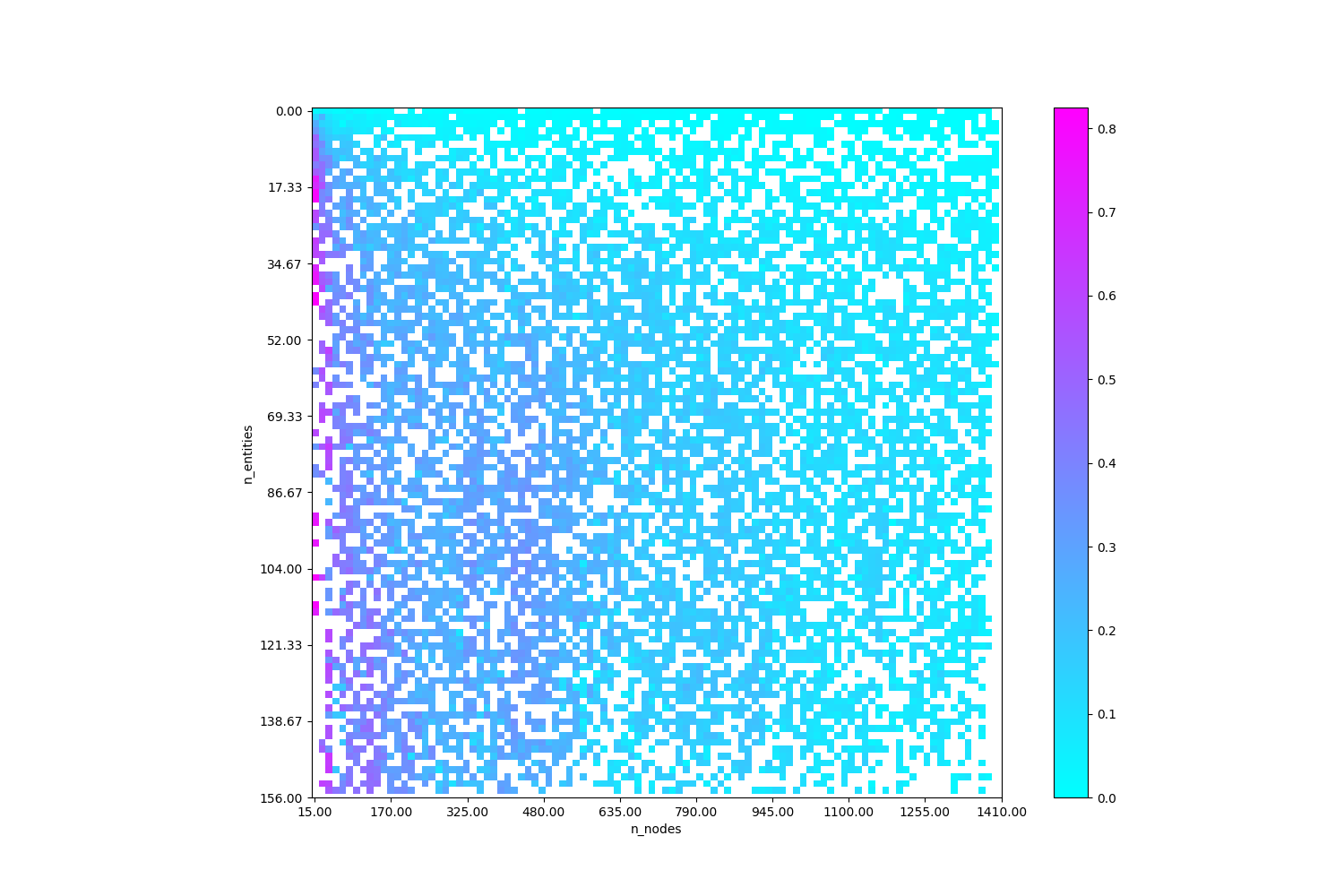}
\caption{Archive after re-evaluation with new random seeds and longer episodes.}
\label{fig:exp1_trial20_string}
\end{subfigure}
\hfill
\caption{Using the archive from a single trial of QD search (left) (with n. entities and n. nodes as BCs), we observe that \textbf{re-evaluating on new seeds} (middle) and with \textbf{longer episodes} (right) leads to increasingly ``holey'' archives, due to random variation in the number of surviving entities after each episode.}
\label{fig:exp1_trial20_reevals}
\hfill
\end{figure}

Because our domain is stochastic---in particular, re-simulating the same fortress (with the same entity class FSMs and initial entity instances and starting positions) will result in different random movement actions from any agent in a `move', `push', or `chase' state---we re-evaluate the the fortresses in the archive using new random seeds and re-insert fortresses into a fresh archive. The results of these re-evaluations (for a single trial) are visualized against the archive resulting from QD search in Figure~\ref{fig:exp1_trial20_reevals}.
In Table 3, we repeat the re-evaluation process for 10 archives, each generated by a separate QD search. We then consider the aggregate archive of overall best elites before and after re-evaluation.

\end{document}